\documentclass[10pt,a4paper,twocolumn]{article}
\usepackage[margin=1.8cm,columnsep=0.6cm]{geometry}
\usepackage[T1]{fontenc}
\usepackage[utf8]{inputenc}
\usepackage{mathptmx}
\usepackage{setspace}
\usepackage{amsmath,amssymb}
\usepackage{graphicx}
\usepackage{booktabs}
\usepackage{tabularx}
\usepackage{array}
\usepackage{xcolor}
\usepackage[numbers,sort&compress]{natbib}
\usepackage{xurl}
\usepackage{lineno}
\usepackage[hidelinks]{hyperref}
\newcommand{\safegraphic}[2]{%
	\IfFileExists{#2}{\includegraphics[#1]{#2}}{%
		\typeout{!!! MISSING FIGURE FILE: #2 !!!}%
		\fbox{\parbox{0.6\linewidth}{\centering\color{red}\small missing figure file:\\ \detokenize{#2}\vspace{1.2cm}}}}}

\title{\bfseries Audit Before You Update: Recording Artifacts and the Limits of Bounded Model Updates in MIMIC-IV ICU Prediction}
\author{Fatema Ferdous Tamanna$^{1,*}$, K.~M.~Merajul Arefin$^{2}$, Md.~Abdul Masud$^{1}$\\[0.5em]
	\small $^{1}$Department of Computer Science and IT, Patuakhali Science and Technology University, Patuakhali, Bangladesh\\
	\small $^{2}$Department of Computer Science and Engineering, University of Dhaka, Dhaka, Bangladesh\\
	\small $^{*}$\texttt{tamanna16@cse.pstu.ac.bd}}
\date{}

\usepackage[labelfont=bf,labelsep=space]{caption}
\begin{document}
	\raggedbottom
	\makeatletter
	\twocolumn[\begin{@twocolumnfalse}
		\maketitle
		\vspace{-1.5em}
	\end{@twocolumnfalse}]
	\makeatother

	\begin{abstract}
		Clinical prediction models in intensive care degrade as data shift, and regulators such as the FDA ask developers to specify bounded update protocols in advance. Evaluations of such protocols assume that apparent drift reflects clinical practice rather than changes in how the database records care. We test whether update strategies can be evaluated on unaudited ICU data, and whether a parameter-bounded update gives up accuracy or attribution stability relative to full fine-tuning once the data are audited. We audit MIMIC-IV 3.1 (2008--2022) for recording artifacts and define a cleaning protocol. On the cleaned cohort, a label-free classifier two-sample test gates updates across three eras. We compare a bounded update, which retrains only treatment-context and fusion parameters, with full fine-tuning for two model families and two targets under a pre-specified non-inferiority margin of 0.015 AUROC, and measure attribution stability against the source model. In 2020--2022 about half of ICU stays have no input or procedure-event records, and a vasopressor ItemID migration makes the 2017--2019 label rate appear to fall when it actually rises. On the cleaned cohort, the gate detects drift in every era, including in physiology-only features, with no false alarms on null data. Bounded updating is non-inferior or superior in 10 of 12 primary cells, but this depends on the target, the model family and learning-rate tuning, and adaptation gains over never updating are small. At a shared learning rate, bounded updates keep physiological attributions more stable in neural models, clearly for vasopressor and less clearly for intubation; with separately tuned rates, full updates are more stable. Audits should precede update evaluation, and the update boundary should be chosen per target.
	\end{abstract}
	{\let\thefootnote\relax\footnotetext{An AI assistant was used to help edit the manuscript; all results were generated by the authors' code and checked by the authors.}}

	\section{Introduction}
	Clinical machine learning models deployed in intensive care units (ICUs) suffer performance degradation over time due to temporal distribution shift driven by evolving clinical guidelines, changing documentation practices, and shifting patient acuity \citep{davis2017calibration,nestor2019feature,guo2022evaluation,finlayson2021clinician}. To ensure patient safety, deployed models must be monitored and periodically updated \citep{jenkins2021continual,feng2022clinical}. However, unconstrained retraining can introduce unpredictable changes in feature attributions, complicating clinical trust and regulatory oversight.
	
	In response, regulatory bodies such as the U.S.\ Food and Drug Administration (FDA) have introduced guidelines for Predetermined Change Control Plans (PCCP) \citep{fda2024pccp}. The PCCP guidance asks AI/ML medical device developers to specify in advance which model components an update will modify, the scope of parameter changes, and the verification metrics used to confirm post-update safety. A natural change-control strategy for multi-modal clinical models is to bound adaptation by semantic data stream: retrain the treatment-context sub-network (which reflects institutional practice changes) while keeping the physiological sequence encoder bit-for-bit immutable. Freezing physiological representations guarantees that patient vitals and laboratory values map to internal representations identically before and after an update.
	
	Any evaluation of such a strategy rests on the data it is run on. Drift detection and update comparison both assume that apparent shifts in features and labels reflect clinical practice, but electronic health record databases also change how and what they record \citep{agniel2018biases,weiskopf2013methods}. This raises two questions: \emph{(i) are update comparisons on a public ICU database valid before the database has been audited for recording artifacts, and (ii) once it has, does a parameter-bounded update give up discriminative accuracy compared to full fine-tuning under a formal non-inferiority margin, and does it deliver the attribution stability the freeze is intended to guarantee?}
	
\begin{figure}[t]
		\centering
		\includegraphics[width=0.42\textwidth]{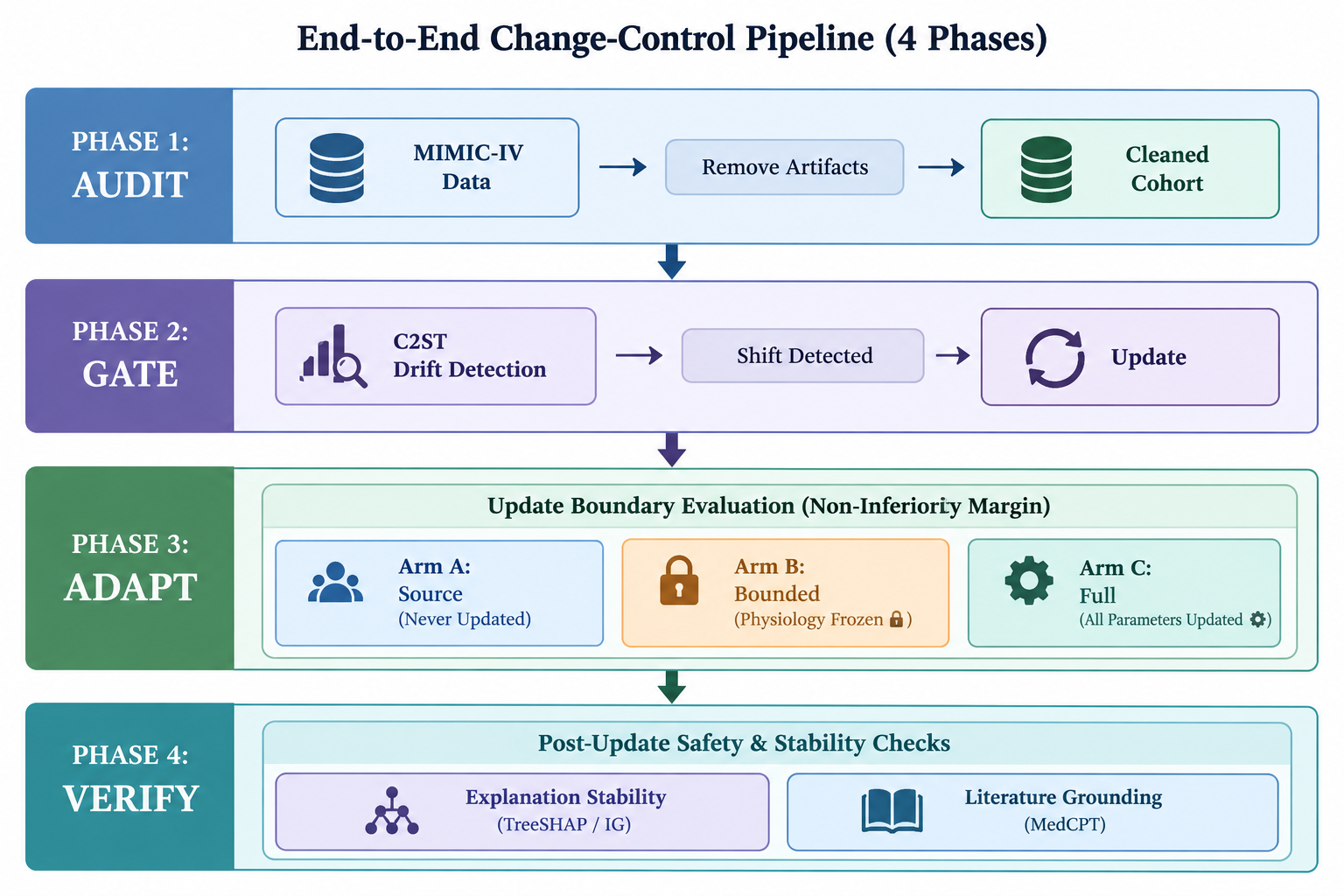}
		\caption{Four-phase change-control protocol used in this paper: pre-monitoring data audit (Phase 1), label-free drift gate (Phase 2), update-boundary evaluation (Phase 3), and post-update verification (Phase 4). In Phase 3, arm A is the never-updated source model, arm B the bounded update (physiology frozen) and arm C the full update.}
		\label{fig:pipeline}
	\end{figure}

	\paragraph{Our Contribution.}
	We argue that an update strategy cannot be judged until the data it is judged on have been audited, and we demonstrate this on MIMIC-IV. The evaluation follows a four-phase change-control protocol (Figure~\ref{fig:pipeline}) in which the audit comes first. Our findings are:
	\begin{itemize}\setlength{\itemsep}{1pt}
		\item \textbf{Recording artifacts distort drift and update evaluation.} A temporal-validity audit of MIMIC-IV 3.1 quantifies a large absence of ICU input- and procedure-event records in 2020--2022 and a vasopressor ItemID migration from 2017, and documents near-absent eMAR records before 2014 and a central-line code change. eMAR availability is documented by MIMIC-IV itself. We document a cleaning protocol and show how the first two change the cohort and the vasopressor label (Section~\ref{sec:res_audit}).
		\item \textbf{On the cleaned data, drift is detectable without labels.} A label-free classifier two-sample test with subject-clustered bootstrapping indicates drift in every era, including in physiology-only features, and gave no false alarms on null data (Section~\ref{sec:res_gate}).
		\item \textbf{Adaptation gains are small, and whether bounded updating gives up accuracy depends on the target, the model family and the tuning protocol.} Gains of full updating over never updating are small (pooled across eras $-0.002$ to $+0.020$ AUROC). Under a pre-specified 0.015 AUROC margin, bounded updating is non-inferior in 10 of 12 primary cells at a shared learning rate. As a secondary finding, at the shared learning rate it gives higher physiological attribution stability in neural models, clearly for vasopressor and less clearly for intubation, where the final-era difference is within seed variation (Sections~\ref{sec:res_disc}--\ref{sec:res_hp}).
	\end{itemize}
	The scientific claims are therefore that audits should precede update evaluation, and that the update boundary should be chosen per target rather than assumed.
	
	\section{Related Work}
	We organize prior work in the order of our protocol: data quality, temporal shift, label-free shift detection, model updating, and attribution stability.
	
	\paragraph{EHR Data Quality and Recording Artifacts.}
	Secondary use of electronic health records depends on data quality, commonly described in terms of completeness, conformance, plausibility, correctness and currency \citep{weiskopf2013methods,kahn2016harmonized}. Records also reflect how care is delivered and documented: whether and when a laboratory test was ordered can predict survival independently of its result \citep{agniel2018biases}. The MIMIC databases document known changes in their sources; MIMIC-III, for example, combines two bedside systems (CareVue and MetaVision) whose fluid-intake records could not be merged in the same way \citep{johnson2016mimiciii}, and MIMIC-IV documents the staged roll-out of eMAR \citep{johnson2023mimic}. Widely used extraction pipelines and benchmarks for MIMIC-III and MIMIC-IV standardize cohort definitions, item mappings and tasks \citep{harutyunyan2019multitask,wang2020mimicextract,gupta2022extensive}. Our audit is complementary: it asks whether table coverage and item definitions are stable across the eras used for drift and update evaluation, and quantifies the consequences when they are not.
	
	\paragraph{Temporal Dataset Shift in Healthcare.}
	Performance degradation in clinical prediction models over time is well documented \citep{davis2017calibration,nestor2019feature,guo2022evaluation}. Nestor et al.\ showed that year-agnostic splits of MIMIC-III hide this degradation and that aggregating raw features into clinical concepts reduces it \citep{nestor2019feature}; Guo et al.\ evaluated domain generalization and adaptation across MIMIC-IV year groups \citep{guo2022evaluation}; and Zhou et al.\ proposed time-aware backtesting of medical models and found that the best training window differs by dataset \citep{zhou2023evaluating}. Finlayson et al.\ describe the sources of dataset shift for clinicians, from changes in technology and practice to changes in population \citep{finlayson2021clinician}, and Subbaswamy and Saria and Sahiner et al.\ review its causes and remedies \citep{subbaswamy2020development,sahiner2023data}. Fern\'andez-Narro et al.\ characterized temporal shift in MIMIC-IV without labels and related the resulting temporal clusters to model performance \citep{fernandeznarro2025unsupervised}. These studies treat the recorded data as given; we show that on MIMIC-IV part of the apparent shift is produced by the recording process.
	
	\paragraph{Label-Free Shift Detection.}
	Detecting dataset shift before outcomes are observed is essential for change-control gating. Classifier two-sample tests (C2ST) \citep{lopez2017c2st} and black-box shift detection (BBSD) \citep{rabanser2019failingloudly} offer formal statistical tests for distribution shift. Applying C2ST to EHR data requires accounting for patient readmissions via grouped folding to avoid false positives.
	
	\paragraph{Model Updating and Regulatory Change Control.}
	Clinical prediction research has long distinguished updating strategies of increasing extent, from recalibration to full model revision, and found that parsimonious updates are often preferable unless validation samples are large \citep{steyerberg2004validation}. More recent work argues for continual monitoring and updating as a routine part of deployment \citep{jenkins2021continual,feng2022clinical}. On the regulatory side, the FDA's PCCP guidance \citep{fda2024pccp} asks for update scope to be pre-specified, and Feng et al.\ framed approval of successive model modifications as an online hypothesis-testing problem, showing that policies without error-rate guarantees allow gradual accuracy erosion (``bio-creep'') \citep{feng2021approval}. In machine learning, restricting or regularizing updates limits forgetting and feature distortion: L2-SP anchors fine-tuned weights to their starting values \citep{li2018l2sp}, elastic weight consolidation protects parameters important to earlier tasks \citep{kirkpatrick2017overcoming}, and tuning only selected layers can improve adaptation under distribution shift \citep{kumar2022finetuning,lee2023surgical}. These methods have mainly been evaluated on vision and language benchmarks structured by network depth. We instead bound updates by semantic input stream, freezing the physiological encoder, and evaluate the bound with a pre-specified non-inferiority margin.
	
	\paragraph{Attribution Stability and Literature Grounding.}
	Post-hoc methods such as TreeSHAP \citep{lundberg2020shap} and Integrated Gradients \citep{sundararajan2017axiomatic} assign feature importances to individual predictions. Explanations are expected to be stable, yet existing methods can change substantially under small input perturbations \citep{alvarezmelis2018robustness}, and different methods often disagree on the same model \citep{krishna2024disagreement}. We use a related notion, stability of attributions across model updates relative to the source model. Linking attributions with dense biomedical retrieval models such as MedCPT \citep{jin2023medcpt} allows predictions to be connected to clinical literature; we measure only whether retrieved documents stay the same, not their relevance.
	
	\section{Data, Models and Evaluation Protocol}
	\label{sec:methods}
	
	\subsection{Data and Eras}
	We use MIMIC-IV version 3.1 \citep{johnson2023mimic} (ICU stays from 2008 through 2022). Each stay was assigned an estimated calendar year using anchor-year midpoints and elapsed time, partitioning the database into a source era (2008--2013) and three 3-year adaptation eras: 2014--2016, 2017--2019, and 2020--2022. All phases after the audit use the cleaned cohort. Within each adaptation era, subjects are ordered by the calendar year of their first stay (ties broken at random; MIMIC-IV date-shifts each patient, so year is the finest comparable resolution), the earlier half forms the adaptation cohort (split by subject, roughly 60/20/20, into update training, early stopping and hyperparameter selection) and the later half the held-out evaluation cohort, and each subject keeps one role across all eras, so no subject appears in both. The ordering is chronological only at the level of a subject's first year: stays of adaptation subjects run to the end of each era, and 70\%, 72\% and 90\% of evaluation stays (5{,}813 of 8{,}328; 6{,}852 of 9{,}470; 2{,}779 of 3{,}101) fall in a year before the last adaptation year.
	
	\subsection{Temporal-Validity Audit (Phase 1)}
	\label{sec:audit_method}
	The audit checks whether the tables and item definitions that feed features and labels are stable across eras, focusing on the completeness and conformance dimensions of EHR data quality \citep{weiskopf2013methods,kahn2016harmonized}. It has four parts.
	\begin{enumerate}\setlength{\itemsep}{1pt}
		\item \textbf{Table coverage.} For each era we compute the share of ICU stays with no \texttt{inputevents} rows and with neither \texttt{inputevents} nor \texttt{procedureevents} rows, both by reconstructed year and by MIMIC's raw anchor-year group.
		\item \textbf{Structure of missingness.} For stays lacking these records we test whether missingness clusters within hospitalizations and patients (multi-stay admissions that are entirely covered, entirely missing, or mixed), and whether it varies by care unit or length-of-stay bin. We compare in-hospital mortality, age and ICU length of stay between stays with and without records, and report standardized mean differences (SMD) for retained versus excluded stays in the modelling cohort.
		\item \textbf{Item definitions.} For each vasoactive agent we count stays per era under the original and an extended ItemID set, and compare the resulting vasopressor label rates. We also check other treatment items used as features, which identified the central-line code change.
		\item \textbf{Documented source changes.} We review the MIMIC-IV documentation for system roll-outs that affect feature availability, which identified eMAR.
	\end{enumerate}
	Each artifact found is either corrected (remapped), excluded (stays or features dropped), or documented as a residual limitation. Results and the resulting cleaning protocol are reported in Section~\ref{sec:res_audit}.
	
	\subsection{Outcomes and Incident Cohorts}
	\label{sec:outcomes}
	Stays of at least 14 hours were included. Features were taken from hours 0--5; outcomes from hours 8--14 (2-hour buffer). Primary endpoints:
	\begin{enumerate}\setlength{\itemsep}{1pt}
		\item \textbf{Incident vasopressor use}: any infusion of a standard vasoactive agent (extended ItemID set) overlapping the outcome window, among stays with none in hours 0--5.
		\item \textbf{Incident invasive ventilation}: any ventilation procedure record or charted ventilator settings (PEEP/tidal volume) overlapping the outcome window, among stays not ventilated in hours 0--5.
	\end{enumerate}
	Evaluations are restricted to incident cohorts (stays already receiving the intervention in hours 0--5 are excluded). Because labels are defined by overlap with hours 8--14, a start during the 6--8 hour buffer also counts as positive. In-hospital mortality (a whole-stay outcome, no incident restriction) is a secondary outcome reported in Appendix~\ref{sec:app_mortality}. A third interventional target, septic shock, was excluded because it is a strict subset of vasopressor use and has very few incident events. Vasopressor use and invasive ventilation are the two primary targets, giving 12 primary cells (2 targets $\times$ 2 model families $\times$ 3 eras).
	
	\subsection{Model Families and Input Streams}
	We evaluate two architectures that decompose inputs into a physiological sequence stream and a treatment-context stream:
	\begin{itemize}\setlength{\itemsep}{1pt}
		\item \textbf{Two-stream neural model.} A 2-layer LSTM (hidden size 64) over 51 physiological channels $\times$ 6 hourly bins; a 2-layer MLP over 49 stay-level treatment features; a fusion head to task logits.
		\item \textbf{Modular gradient-boosted model.} Separate XGBoost boosters on flattened physiology and on treatment features, combined by a logistic-regression fusion head.
	\end{itemize}
	Observation masks and related care-process signals are aggregated to stay-level rates and assigned to the treatment stream. Note that some channels conventionally labelled ``physiology'' (e.g., invasive pressures, GCS components, laboratory availability) still reflect care processes to a non-negligible degree.
	
	\subsection{Label-Free Drift Gate}
	\label{sec:gate_method}
	To assess the need for an update without waiting for delayed outcome labels, we use a classifier two-sample test (C2ST) \citep{lopez2017c2st}. At adaptation step $k$, a histogram gradient-boosting classifier discriminates the reference feature distribution (source training data plus the training parts of earlier adaptation cohorts) from the training part of the new era's adaptation cohort. Each side of a test is subsampled to 2,000 stays. Folds use Stratified Group K-Fold by patient ID so that readmissions never straddle train and test. We compute 5-fold out-of-fold AUROC and a 300-resample subject-cluster bootstrap. The gate indicates an update if the lower 95\% bound of that AUROC (the 2.5th percentile of the bootstrap) exceeds 0.60:
	\begin{equation}
		\text{Gate Decision} = \mathbb{I}\bigl(\text{C2ST\_AUROC}_{95\%\text{ lower bound}} > 0.60\bigr).
	\end{equation}
	On null data (subject-disjoint halves of the source period) the rule yielded a 0\% false-alarm rate (0/40 trials; mean AUROC 0.4996). Adaptation was run in every era so that both update strategies could be evaluated, and the gate was replayed afterwards (Section~\ref{sec:limitations}).
	
	\subsection{Update Arms}
	Starting from a source model trained on 2008--2013 data, the primary comparison is:
	\begin{itemize}\setlength{\itemsep}{1pt}
		\item \textbf{A (Source).} Never updated; evaluated statically on later eras.
		\item \textbf{B (Bounded).} Neural: retrain only the treatment branch and fusion head ($\approx$17.8\% of parameters); the physiology encoder remains bit-for-bit frozen. Trees: fit a new booster on the treatment features and refit the fusion head; the existing physiology booster is untouched.
		\item \textbf{C (Full).} Neural: retrain all parameters. Trees: fit new boosters on both branches (existing trees are not modified) and refit the fusion head.
	\end{itemize}
	Each update uses the training part of the era's adaptation cohort plus a replay sample (up to 500 stays drawn from earlier eras' adaptation cohorts; none at the first update), with early stopping on held-out adaptation stays. Both B and C share a single hyperparameter setting per architecture (neural learning rate $3\times10^{-3}$ for the two primary targets and $3\times10^{-4}$ for mortality; XGBoost $\eta = 0.03$). When an arm's rate was selected in its own sweep, the sweep was run once, at the first update, on the selection split, and the chosen value was then held for later updates. The $3\times10^{-3}$ rate is the one B selected in its own sweep; it is the largest value in the grid ($3\times10^{-5}$ to $3\times10^{-3}$), so B's optimum may lie beyond it. C's own sweep selected $3\times10^{-4}$, and the sensitivity to this choice is reported in Section~\ref{sec:res_hp}. Loss, optimizer and tree settings are given in the released code. The conceptual difference between arms B and C is illustrated in Appendix~\ref{sec:app_arch}.
	
	\subsection{Non-Inferiority Protocol}
	\label{sec:ni}
	The pre-specified non-inferiority margin is 0.015 AUROC; we follow standard non-inferiority reporting practice by stating the margin, the confidence level used for the decision, and the intervals \citep{piaggio2012reporting}. Discrimination is measured for the models obtained after the third update, on each era's evaluation cohort. For each era, label, and model family we report seed-averaged incident AUROC (5 seeds) and paired stay-level bootstrap confidence intervals (1,000 resamples) for B$-$C and C$-$A. Bounded update B is declared non-inferior to full update C when the lower end of the 90\% bootstrap interval of B$-$C (its 5th percentile, a one-sided test at $\alpha = 0.05$) exceeds $-0.015$; ``sup.'' means the lower end of the 95\% interval exceeds 0. Table~\ref{tab:accuracy} prints 95\% intervals alongside the verdict from the 90\% interval. Pooled contrasts across eras (post-hoc) use event-weighted means and a shared subject-cluster bootstrap.
	
	\subsection{Post-Update Verification}
	Feature attributions are obtained with TreeSHAP (trees) and Integrated Gradients (neural models; 50 steps, training-mean baseline). Attribution-conditioned PubMed retrieval uses MedCPT \citep{jin2023medcpt}: the top-5 physiological and top-4 treatment features by absolute attribution form short queries that are matched against a fixed 2009--2019 abstract corpus. Stability is measured by the per-patient Spearman correlation of signed attributions (within each stream and over all 100 inputs; neural physiology attributions are summed over the six hours) and by Jaccard overlap of top retrieved documents, both relative to the source model. We call these measures \emph{attribution stability} and \emph{retrieval stability}: they quantify how much attributions and retrieved documents move, not whether either is clinically correct (retrieval relevance was not rated).
	
	\section{Results}
	
	\subsection{Phase 1: Recording Artifacts in MIMIC-IV}
	\label{sec:res_audit}
	
	Table~\ref{tab:audit} summarizes the principal artifacts, their magnitude, and the cleaning protocol applied before any drift detection or adaptation.
	
	\begin{table*}[t]
		\centering
		\caption{MIMIC-IV temporal-validity audit: principal recording artifacts and cleaning protocol. All subsequent results are reported on the cleaned cohort.}
		\label{tab:audit}
		\small
		\begin{tabularx}{\textwidth}{l p{1.6cm} >{\raggedright\arraybackslash}p{5.4cm} >{\raggedright\arraybackslash}X}
			\toprule
			\textbf{Artifact} & \textbf{Eras} & \textbf{Impact} & \textbf{Cleaning protocol} \\
			\midrule
			Missing ICU tables & 2020--2022 & 53.3\% of stays (8,933 of 16,745) lack any \texttt{inputevents} rows; 53.2\% lack both \texttt{inputevents} and \texttt{procedureevents}. This bin includes 761 stays whose reconstructed year is after 2022, all of them missing; excluding them, 51.0\% (8,147 of 15,984) lack both. & Exclude stays with neither \texttt{inputevents} nor \texttt{procedureevents} rows (below 1\% in each earlier era, by reconstructed year); treatment features are unobserved. \\
			\addlinespace
			Vasopressor ItemIDs & 2017--2022 & Phenylephrine (and related) preparations reassigned new ItemIDs. Counting 229630 and 229632 as well as 221749 raises the number of stays with phenylephrine from 1,845 to 3,495 (+89\%) in 2017--2019 and from 311 to 1,147 (+269\%) in 2020--2022, versus +6 to +8\% in 2008--2016 (counts by MIMIC anchor-year group). & Remap to a unified extended ItemID set (adding 229617 and 229630--229632) across all eras. \\
			\addlinespace
			eMAR absence & Pre-2014 & Medication-administration records near-absent before rollout in 2011--2013, as documented by MIMIC-IV (coverage incomplete into 2014--2016). & Exclude all eMAR-derived features from model inputs. \\
			\addlinespace
			Central-line codes & 2017--2022 & Cordis code (224267) replaced by MAC/Single-Lumen codes. & Documented residual undercount; not fully remapped in the primary cohort. \\
			\bottomrule
		\end{tabularx}
	\end{table*}
	
	\paragraph{Missing ICU tables.} Excluding stays that lack ICU input/procedure records was necessary because their treatment features and vasopressor labels are structurally empty. The counts below use MIMIC's raw anchor-year group (10,797 stays in 2020--2022), so they differ from the reconstructed-year figures in Table~\ref{tab:audit} (53.2\% of 16,745 stays). In this group 6,272 stays (58.1\%) have no records, and the gap clusters at the hospitalization level: of 1,019 admissions with two or more ICU stays, 400 were entirely covered, 591 entirely missing, and only 28 (2.7\%) mixed; at the patient level 167 of 1,529 (10.9\%) were mixed. The gap is spread across care units (55--60\% in every unit with at least 400 stays) and length-of-stay bins (56--61\%). This pattern is consistent with systematic administrative boundaries rather than random non-reporting. Stays with and without records were similar: in-hospital mortality was 14.0\% with records versus 14.1\% without, mean age 63.8 versus 63.9 years, and median ICU stay 2.15 versus 2.33 days. In the modelling cohort, age, weight, sex and in-hospital mortality were likewise balanced between retained and excluded stays (age standardized mean difference [SMD] $= 0.000$, female sex 0.007, mortality 0.027; Appendix~\ref{sec:app_balance}). This balance does not rule out selection on unmeasured variables.
	
	\paragraph{Vasopressor ItemIDs.} Phenylephrine infusions are recorded under new ItemIDs from 2017 onward (Table~\ref{tab:audit}). Without the extended ItemID set, a stay whose only vasopressor records use the new phenylephrine ItemIDs is labelled negative. In the cleaned data the extended set raises the vasopressor label rate from 20.5\% to 23.9\% in 2017--2019 and from 22.5\% to 27.8\% in 2020--2022, with no change before 2014 and 0.01 percentage points in 2014--2016; the old set would show an apparent drop in 2017--2019 (from 22.0\% in 2014--2016 to 20.5\%) where the rate actually rises (to 23.9\%). The central-line change remains a documented residual.
	
	\paragraph{eMAR.} MIMIC-IV documents that the eMAR system was implemented during 2011--2013 and is not available for all patients\footnote{\url{https://mimic.mit.edu/docs/iv/modules/hosp/emar/}}, so eMAR-derived features are near zero in the source era; we exclude them. (In an earlier pipeline version that kept them, setting them to their source-era value changed the never-updated neural model's AUROC by $+0.039$ to $+0.068$.)
	
	\subsection{Phase 2: Label-Free Drift Gate}
	\label{sec:res_gate}
	
	Table~\ref{tab:drift} reports the C2ST scores and BBSD $p$-values. The gate indicated updates in every era (lower 95\% bounds 0.702, 0.787 and 0.755, all above 0.60); the two source-period year groups, by comparison, gave AUROC 0.593 with lower bound 0.578. Physiology-only C2ST also exceeded the cutoff in every era (0.685--0.747), indicating that physiological sequence distributions shifted along with treatment context; treatment-only AUROC in 2014--2016 (0.597) did not exceed it. BBSD flagged the tree model's own outputs in 2017--2019 ($p = 0.009$) and the neural model in 2020--2022 ($p < 1\times10^{-5}$).
	
	\begin{table}[t]
		\centering
		\caption{Label-free drift detection by era: C2ST AUROC with lower 95\% bound in parentheses (all, physiology-only, treatment-only features), BBSD $p$-values (Bonferroni across labels), and whether the update was indicated (lower bound $> 0.60$).}
		\label{tab:drift}
		\small
		\resizebox{\columnwidth}{!}{\begin{tabular}{lcccccc}
				\toprule
				Era & C2ST all & C2ST physiology & C2ST treatment & BBSD $p$ (tree) & BBSD $p$ (NN) & Drift \\
				\midrule
				2014 - 2016 & 0.715 (0.702) & 0.699 (0.685) & 0.597 (0.578) & 0.30 & 0.30 & yes \\
				2017 - 2019 & 0.799 (0.787) & 0.761 (0.747) & 0.720 (0.706) & 9e-03 & 0.15 & yes \\
				2020 - 2022 & 0.770 (0.755) & 0.728 (0.714) & 0.728 (0.713) & 0.42 & 5e-06 & yes \\
				\bottomrule
		\end{tabular}}
	\end{table}
	
	\subsection{Phase 3: Bounded versus Full Update}
	\label{sec:res_disc}
	
\begin{figure*}[t]
		\centering
		\includegraphics[width=0.95\textwidth]{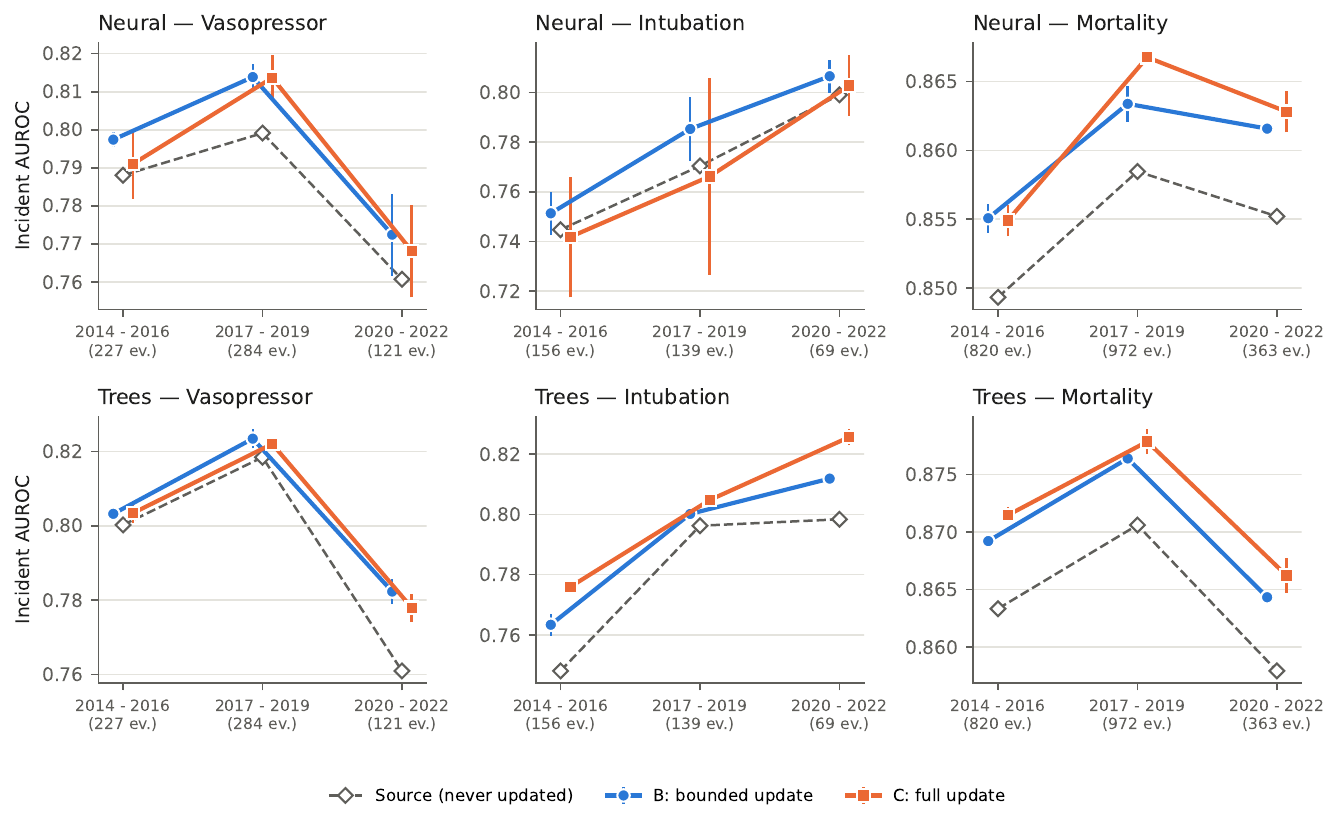}
		\caption{Incident AUROC of the final models on each era's evaluation cohort. Points are 5-seed means; error bars are seed standard deviations. Top: two-stream neural model. Bottom: modular gradient-boosted trees. Columns are vasopressor, intubation and, as a secondary outcome, mortality (Appendix~\ref{sec:app_mortality}). Primary comparison is bounded update (B) versus full update (C); source model (A) is shown as reference. Both arms share a single hyperparameter setting per architecture (neural learning rate $3\times10^{-3}$ for vaso/intubation and $3\times10^{-4}$ for mortality; XGBoost $\eta = 0.03$).}
		\label{fig:auroc}
	\end{figure*}

	Figure~\ref{fig:auroc} and Table~\ref{tab:accuracy} summarize incident AUROC for the two primary targets. At the shared hyperparameters, bounded updating was non-inferior or superior to full updating in 10 of 12 primary evaluation cells.
	
	\begin{table*}[t]
		\centering
		\caption{Incident AUROC (mean of 5 seeds) and paired differences for the two primary targets. Margin $= 0.015$ AUROC. Primary comparison: Bounded (B) vs Full (C); Source (A) as reference. Both arms share identical hyperparameters. The source model is a single model per family; B and C are 5-seed means. Intervals are paired stay-level bootstrap 95\% CIs; the NI verdict uses the lower end of the 90\% interval against $-0.015$ (``--'': non-inferiority not shown; ``sup.'': lower end of the 95\% interval above 0).}
		\label{tab:accuracy}
		\scriptsize
		\resizebox{\textwidth}{!}{\begin{tabular}{lllrrrrrll}
				\toprule
				Label & Model & Era & Events & Source & B & C & C$-$Source [95\% CI] & B$-$C [95\% CI] & \\
				\midrule
				Vasopressor & Neural & 2014 - 2016 & 227 & 0.788 & 0.797$\pm$0.002 & 0.791$\pm$0.009 & +0.003 [-0.013, +0.020] & +0.006 [-0.004, +0.016] & NI \\
				Vasopressor & Neural & 2017 - 2019 & 284 & 0.799 & 0.814$\pm$0.003 & 0.814$\pm$0.006 & +0.015 [-0.002, +0.032] & 0.000 [-0.009, +0.009] & NI \\
				Vasopressor & Neural & 2020 - 2022 & 121 & 0.761 & 0.772$\pm$0.011 & 0.768$\pm$0.012 & +0.007 [-0.022, +0.037] & +0.004 [-0.008, +0.017] & NI \\
				\addlinespace
				Vasopressor & Trees & 2014 - 2016 & 227 & 0.800 & 0.803$\pm$0.001 & 0.803$\pm$0.003 & +0.003 [-0.009, +0.016] & 0.000 [-0.007, +0.007] & NI \\
				Vasopressor & Trees & 2017 - 2019 & 284 & 0.818 & 0.823$\pm$0.003 & 0.822$\pm$0.001 & +0.004 [-0.008, +0.015] & +0.001 [-0.005, +0.007] & NI \\
				Vasopressor & Trees & 2020 - 2022 & 121 & 0.761 & 0.782$\pm$0.003 & 0.778$\pm$0.004 & +0.017 [-0.005, +0.037] & +0.004 [-0.005, +0.015] & NI \\
				\addlinespace
				Intubation & Neural & 2014 - 2016 & 156 & 0.745 & 0.751$\pm$0.009 & 0.742$\pm$0.024 & -0.003 [-0.027, +0.022] & +0.010 [-0.009, +0.029] & NI \\
				Intubation & Neural & 2017 - 2019 & 139 & 0.770 & 0.785$\pm$0.013 & 0.766$\pm$0.040 & -0.004 [-0.027, +0.019] & +0.019 [+0.006, +0.034] & sup. \\
				Intubation & Neural & 2020 - 2022 & 69 & 0.799 & 0.807$\pm$0.007 & 0.803$\pm$0.012 & +0.004 [-0.026, +0.034] & +0.004 [-0.017, +0.025] & NI \\
				\addlinespace
				Intubation & Trees & 2014 - 2016 & 156 & 0.748 & 0.763$\pm$0.004 & 0.776$\pm$0.001 & +0.028 [+0.007, +0.050] & -0.013 [-0.028, +0.001] & -- \\
				Intubation & Trees & 2017 - 2019 & 139 & 0.796 & 0.800$\pm$0.002 & 0.805$\pm$0.002 & +0.009 [-0.009, +0.025] & -0.005 [-0.017, +0.008] & NI \\
				Intubation & Trees & 2020 - 2022 & 69 & 0.798 & 0.812$\pm$0.001 & 0.826$\pm$0.003 & +0.027 [-0.004, +0.060] & -0.014 [-0.032, +0.004] & -- \\
				\bottomrule
		\end{tabular}}
	\end{table*}
	
	\paragraph{Vasopressor use.} Bounded and full updates were essentially equivalent across eras and families (pooled B$-$C $+0.002$ AUROC, 95\% CI $-0.001$ to $+0.006$; Table~\ref{tab:pooled}).
	
	\paragraph{Invasive ventilation.} Results depended on architecture. In neural models, bounded update B was non-inferior or superior in all three eras and better in pooled contrast ($+0.012$, 95\% CI $+0.002$ to $+0.022$). In tree models, full update C outperformed B by 0.005--0.014 AUROC per era (pooled $-0.010$, $-0.018$ to $-0.001$). The advantage of full updating is therefore target- and architecture-dependent rather than uniform.
	
	\begin{table}[t]
		\centering
		\caption{Pooled AUROC contrasts across eras (event-weighted; subject-clustered 95\% CIs; post-hoc). Combined is the unweighted mean of the two families.}
		\label{tab:pooled}
		\small
		\resizebox{\columnwidth}{!}{\begin{tabular}{llll}
				\toprule
				Target & Family & Bounded vs Full ($B - C$) & Full vs Source ($C - A$) \\
				\midrule
				Vasopressor & Neural & $+0.003$ [$-0.003$, $+0.009$] & $+0.009$ [$-0.002$, $+0.021$] \\
				Vasopressor & Trees  & $+0.001$ [$-0.003$, $+0.006$] & $+0.006$ [$-0.001$, $+0.014$] \\
				Vasopressor & Combined & $+0.002$ [$-0.001$, $+0.006$] & $+0.008$ [$0.000$, $+0.016$] \\
				\addlinespace
				Intubation & Neural & $+0.012$ [$+0.002$, $+0.022$] & $-0.002$ [$-0.017$, $+0.013$] \\
				Intubation & Trees  & $-0.010$ [$-0.018$, $-0.001$] & $+0.020$ [$+0.007$, $+0.033$] \\
				Intubation & Combined & $+0.001$ [$-0.006$, $+0.008$] & $+0.009$ [$-0.002$, $+0.020$] \\
				\bottomrule
		\end{tabular}}
	\end{table}
	
	\paragraph{Adaptation gains are small.} Gains of the full update over the source model were modest (per cell $-0.004$ to $+0.028$ AUROC; pooled across eras $-0.002$ to $+0.020$ depending on target and family). The practical value of \emph{any} update protocol on these tasks is therefore limited. The never-updated source model (A) already satisfied non-inferiority versus full update C in 2 of the 12 primary cells (both vasopressor, trees) and in all 6 mortality cells, which underlines how small the absolute gains are and how little the 0.015 margin discriminates.
	
	\subsection{Phase 4: Attribution and Retrieval Stability}
	\label{sec:res_expl}
	
	In the final era (2020--2022), the share of attribution change that fell on treatment features for the bounded neural update was 60.1\% (vasopressor) and 63.0\% (intubation); for the full update it was 47.7\% and 63.4\%. The bounded update therefore concentrated change on treatment features for vasopressor but not distinctly for intubation. The fusion head is trainable in B, so physiological attributions can still move. Table~\ref{tab:explanation} reports Spearman rank correlation of per-patient feature attributions against the source model.
	
	\begin{table}[t]
		\centering
		\caption{Attribution stability in the final era (2020--2022): mean Spearman rank correlation $\rho$ of per-patient feature attributions against the source model (5-seed means), for the two primary targets. Neural models use the shared learning rate of $3\times10^{-3}$. Tree physiology $= 1.000$ by construction (rank-identical).}
		\label{tab:explanation}
		\small
		\resizebox{\columnwidth}{!}{\begin{tabular}{llrcccccc}
				\toprule
				& & & \multicolumn{2}{c}{Physiology} & \multicolumn{2}{c}{Treatment} & \multicolumn{2}{c}{All} \\
				Label & Model & $n$ & B & C & B & C & B & C \\
				\midrule
				Vasopressor & Neural & 2124 & 0.635 & 0.438 & 0.418 & 0.547 & 0.513 & 0.499 \\
				Vasopressor & Trees & 2124 & 1.000 & 0.780 & 0.462 & 0.468 & 0.763 & 0.652 \\
				Intubation & Neural & 2044 & 0.503 & 0.455 & 0.699 & 0.748 & 0.608 & 0.618 \\
				Intubation & Trees & 2044 & 1.000 & 0.680 & 0.762 & 0.765 & 0.838 & 0.722 \\
				\bottomrule
		\end{tabular}}
	\end{table}
	
	\paragraph{Neural models.} At the shared rate of $3\times10^{-3}$ (other rate protocols are in Section~\ref{sec:res_hp}), for vasopressor prediction, bounded update B showed higher physiological attribution stability than full update C ($0.635$ vs $0.438$), with B higher in 4 of 5 seeds. For intubation the final-era difference is small ($0.503$ vs $0.455$): B is higher in 3 of 5 seeds and the difference is within seed-to-seed variation (seed SD 0.14). In the first two eras B was higher in all 5 seeds for both targets, so the intubation evidence is consistent in direction but weakest in the final era. We did not test whether per-seed attribution stability relates to per-seed AUROC change, so we do not claim that attribution stability explains the accuracy results.
	
	\paragraph{Tree models.} Physiological rank correlation is $\rho = 1$ under B by construction (the physiology booster is never updated; the head only rescales it). This is a property of the design and not an empirical result. Under C the corresponding correlations for the two primary targets are 0.68--0.78.
	
	\paragraph{Treatment attributions.} Treatment attributions were generally less stable under B than under C for neural models (vasopressor 0.418 vs 0.547; intubation 0.699 vs 0.748) and similar for trees. This is expected: under B the treatment branch and fusion head are the parts being retrained, so most of the change is routed there. It is a trade-off of bounding, since stability of the physiology stream is bought with larger change in the treatment stream.
	
	\paragraph{Retrieval.} Literature retrieval stability (Jaccard overlap of the top-5 MedCPT documents with the source model; Appendix~\ref{sec:app_retrieval}) shows the same qualitative pattern for physiology queries: queries from bounded-update attributions keep higher overlap than those from full updates. It measures stability only; retrieval relevance was not rated.
	
	\subsection{Hyperparameter Sensitivity}
	\label{sec:res_hp}
	The neural result depends on how learning rates are set. A shared rate isolates the effect of freezing; separately tuned rates compare each strategy at its own best. With a shared rate of $3\times10^{-3}$ (B's own optimum), B is non-inferior or superior in all 6 vasopressor/intubation neural cells. With a shared rate of $3\times10^{-4}$ (C's own optimum), B is non-inferior in 5 of 6. If each arm uses its own validation-selected rate, B is non-inferior in only 2 of 6 neural cells; for trees, 5 of 6 verdicts are unchanged. Bounded updating is therefore non-inferior when both arms train at the same rate, whichever arm's optimum it is, but not when each is tuned separately (Appendix~\ref{sec:app_hp}). The attribution result depends on the rate as well. In the final era, mean physiological attribution stability (B vs C) is 0.635 vs 0.438 for vasopressor and 0.503 vs 0.455 for intubation at the shared rate of $3\times10^{-3}$; 0.885 vs 0.806 and 0.802 vs 0.820 at the shared rate of $3\times10^{-4}$; and 0.635 vs 0.813 and 0.503 vs 0.830 with separately tuned rates. The attribution advantage of the bounded update is therefore not robust to the learning-rate protocol.
	
	\section{Discussion}
	The audit is the part of this study that carries the most weight. In 2020--2022, about half of raw ICU stays (53.2\%; 51.0\% excluding reconstructed years after 2022) have neither \texttt{inputevents} nor \texttt{procedureevents} records (below 1\% in each earlier era, by reconstructed year), a new set of phenylephrine ItemIDs appears from 2017, and eMAR records, as documented by MIMIC-IV, are near-absent before 2014. The ItemID remap raises the vasopressor label rate by 3.4 and 5.3 percentage points in 2017--2019 and 2020--2022; without it, the rate would appear to fall in 2017--2019 although it rises. Without the audit, drift detection and update comparisons would have been contaminated by database changes rather than by clinical practice. The audit is incomplete (the central-line change is not remapped), and the mechanism behind the missing ICU tables is unknown.
	
	On the audited data, the label-free gate indicated updates in all three adaptation eras. The two source-period year groups already give a C2ST AUROC of 0.593 (lower bound 0.578), just below the decision threshold, and physiology-only C2ST also fires in every era: physiological inputs shift too, which weakens the premise behind freezing the physiology encoder. Some detected shift may reflect charting or recording changes rather than clinical practice change. Adaptation gains over never updating are small, so the practical value of any update is limited on these tasks.
	
	Within that setting, bounded updates (treatment-context and fusion parameters only) were non-inferior or superior to full fine-tuning in 10 of 12 primary cells at matched hyperparameters. In neural models at the shared $3\times10^{-3}$ rate, their mean physiological attribution stability was higher than that of full updates in every era for both targets, most clearly for vasopressor; in trees, physiology attributions are rank-identical by construction. This makes bounded updating a workable option for change-control frameworks such as the FDA PCCP, but not a default: full updating was better for tree intubation models, the neural result holds only at a shared learning rate (and the attribution advantage reverses with separately tuned rates), the final-era intubation attribution difference is within seed variation, and adaptation gains were small on both primary targets. The update boundary should be chosen per clinical target.
	
	\section{Limitations}
	\label{sec:limitations}
	\begin{itemize}\setlength{\itemsep}{2pt}
		\item \textbf{Scope.} All results come from MIMIC-IV, a single health system; external validation (e.g., eICU) is needed before clinical claims. Retrieval relevance was not rated, so the retrieval results measure stability only.
		\item \textbf{Residual artifacts.} The central-line ItemID change was not remapped; \texttt{has\_central\_line\_obs} undercounts after 2017, and its effect on the B-versus-C comparison was not quantified.
		\item \textbf{Gate.} Updates were run in every era and the gate replayed afterwards; the 0.60 threshold was chosen after inspecting null-data behaviour; and because the gate fired in every era, the no-update branch was not exercised.
		\item \textbf{Comparison design.} The neural non-inferiority result holds at a shared learning rate (6 of 6 cells at B's optimum, 5 of 6 at C's) but in only 2 of 6 cells with separately tuned rates (Section~\ref{sec:res_hp}); the headline rate $3\times10^{-3}$ was not pre-specified and is the largest value in the learning-rate grid. The 0.015 margin is similar in size to the adaptation gains, and the source model (A) passes the same test in 2 of 12 primary cells and all 6 mortality cells.
		\item \textbf{Evaluation timing.} Per-era AUROC is measured for the final models, not for the model in use during each era, so the comparison is retrospective.
		\item \textbf{Temporal separation.} Adaptation and evaluation cohorts are disjoint by subject but ordered in time only at the first-year level, and most evaluation stays fall within the calendar span of the adaptation cohort (70--90\%), so the adaptation results are not a strict prospective test; this applies to both arms alike.
		\item \textbf{Statistics.} Per-cell bootstrap intervals resample stays, not subjects, and are therefore narrower than subject-clustered intervals would be; pooled intervals are subject-clustered and post hoc. The source model is a single model per family (the median-validation model of five seeds), so the seeds vary only the adaptation.
		\item \textbf{Attribution stability.} Physiological rank stability of 1.0 for trees under B follows from the design, and physiology-only C2ST fires in all three eras, so physiological inputs are not stationary. The neural advantage of B in physiological attribution stability holds at the shared $3\times10^{-3}$ rate and reverses with separately tuned rates (Section~\ref{sec:res_hp}). Stability of attributions does not show that the attributions are correct.
	\end{itemize}
	
	\section{Conclusion}
	Before any ICU model update is evaluated, the data should be audited: on MIMIC-IV, ICU input and procedure records are missing for about half of 2020--2022 stays, and the phenylephrine ItemID migration changes the vasopressor label rate by up to 5.3 percentage points. On the audited data, drift is detectable by a label-free gate that gave no false alarms on null data, but adaptation gains are small. A parameter-bounded update that freezes the physiological encoder was non-inferior to full adaptation on discrimination in 10 of 12 primary cells at matched hyperparameters and, at a shared learning rate, gave more stable physiological attributions in neural models and, by construction, in trees. Full updating was better for tree intubation models, and the neural result holds at a shared learning rate but not with separately tuned rates. The appropriate scope of an update should be determined per clinical target.

	\subsubsection*{Code Availability}
	Code to reproduce every table and figure is present here: https://github.com/empresst/DataAudit.git
	\subsubsection*{Broader Impact Statement}
	This is a retrospective analysis of a de-identified public database and makes no
	recommendation to deploy any model. Bounded updates can make model changes easier
	to declare and verify, but our results show that a bound can cost a large share of
	an update's benefit for some prediction targets; adopting such a bound without a
	per-target evaluation could therefore degrade a deployed model's discrimination.
	
	\subsubsection*{Data Availability and Ethics}
	MIMIC-IV is available to credentialed users through PhysioNet under a data use
	agreement; no patient-level data are shared. The original data collection was
	approved with a waiver of informed consent.

	\appendix
	
	\section{Supplementary Analyses}
	
	\subsection{Hyperparameter Sensitivity and Tuning Parity}
	\label{sec:app_hp}
	In the main experiments, both the bounded (B) and full (C) arms used a single shared learning-rate configuration per architecture. When the full-update arm was instead allowed to select its own best hyperparameter from the search grid, the non-inferiority conclusions were unchanged for the tree models (5 of 6 evaluation cells). For neural architectures, independent optimization shifted 4 of 6 neural comparisons into the inconclusive margin: Arm C's own validation sweep selected a lower learning rate ($3\times10^{-4}$) than Arm B's ($3\times10^{-3}$), so non-inferiority held in only 2 of 6 neural cells. A plausible (untested) explanation is that Arm C, which trains all parameters, benefits from a lower rate than Arm B's compact treatment-plus-fusion block (17.8\% of parameters). With both arms matched at $3\times10^{-4}$, B was non-inferior in 5 of 6 neural cells. The matched design was retained for the primary analysis because it isolates the parameter-freezing choice from tuning differences, but the neural non-inferiority conclusion should be read as conditional on it. Physiology-only and L2-SP-anchored update controls were run only at unequal learning rates and are not reported.
	
	\subsection{Secondary Outcome: In-Hospital Mortality}
	\label{sec:app_mortality}
	In-hospital mortality is a secondary outcome. It is a whole-stay outcome predicted from the first six hours, with no incident restriction, and stays with death before hour 14 are not in the cohort. It is not included in the 12 primary cells. Table~\ref{tab:app_mort_acc} gives discrimination and Tables~\ref{tab:app_mort_stab} and~\ref{tab:app_mort_ret} give attribution and retrieval stability. Bounded updating was non-inferior to full updating in all 6 mortality cells, and the never-updated source model (A) was also non-inferior to full updating in all 6, so the comparison is uninformative about the update boundary for this target.
	
	\begin{table*}[!ht]
		\centering
		\caption{Mortality: AUROC (mean of 5 seeds) and paired differences. Margin $= 0.015$ AUROC. Conventions as in Table~\ref{tab:accuracy}.}
		\label{tab:app_mort_acc}
		\scriptsize
		\resizebox{\textwidth}{!}{\begin{tabular}{lllrrrrrll}
				\toprule
				Label & Model & Era & Events & Source & B & C & C$-$Source [95\% CI] & B$-$C [95\% CI] & \\
				\midrule
				Mortality & Neural & 2014 - 2016 & 820 & 0.849 & 0.855$\pm$0.001 & 0.855$\pm$0.001 & +0.006 [+0.002, +0.009] & 0.000 [-0.002, +0.003] & NI \\
				Mortality & Neural & 2017 - 2019 & 972 & 0.858 & 0.863$\pm$0.001 & 0.867$\pm$0.000 & +0.008 [+0.005, +0.012] & -0.003 [-0.006, -0.001] & NI \\
				Mortality & Neural & 2020 - 2022 & 363 & 0.855 & 0.862$\pm$0.000 & 0.863$\pm$0.001 & +0.008 [+0.002, +0.013] & -0.001 [-0.005, +0.002] & NI \\
				Mortality & Trees & 2014 - 2016 & 820 & 0.863 & 0.869$\pm$0.000 & 0.871$\pm$0.001 & +0.008 [+0.004, +0.012] & -0.002 [-0.005, 0.000] & NI \\
				Mortality & Trees & 2017 - 2019 & 972 & 0.871 & 0.876$\pm$0.001 & 0.878$\pm$0.001 & +0.007 [+0.004, +0.011] & -0.001 [-0.004, +0.001] & NI \\
				Mortality & Trees & 2020 - 2022 & 363 & 0.858 & 0.864$\pm$0.000 & 0.866$\pm$0.001 & +0.008 [+0.003, +0.014] & -0.002 [-0.006, +0.002] & NI \\
				\bottomrule
		\end{tabular}}
	\end{table*}
	
	\begin{table}[!ht]
		\centering
		\caption{Mortality, final era (2020--2022): attribution stability (mean Spearman $\rho$ against the source model).}
		\label{tab:app_mort_stab}
		\small
		\resizebox{\columnwidth}{!}{\begin{tabular}{llrcccccc}
				\toprule
				& & & \multicolumn{2}{c}{Physiology} & \multicolumn{2}{c}{Treatment} & \multicolumn{2}{c}{All} \\
				Label & Model & $n$ & B & C & B & C & B & C \\
				\midrule
				Mortality & Neural & 3101 & 0.964 & 0.893 & 0.920 & 0.963 & 0.926 & 0.932 \\
				Mortality & Trees & 3101 & 1.000 & 0.827 & 0.835 & 0.862 & 0.918 & 0.845 \\
				\bottomrule
		\end{tabular}}
	\end{table}
	
	\begin{table}[!ht]
		\centering
		\caption{Mortality, final era (2020--2022): Jaccard similarity of top-5 MedCPT documents versus the source model (300 cases per label).}
		\label{tab:app_mort_ret}
		\small
		\resizebox{\columnwidth}{!}{\begin{tabular}{llrcccc}
				\toprule
				& & & \multicolumn{2}{c}{Physiology query} & \multicolumn{2}{c}{Treatment query} \\
				Label & Model & Cases & B & C & B & C \\
				\midrule
				Mortality & Neural & 300 & 0.638 & 0.509 & 0.596 & 0.696 \\
				Mortality & Trees & 300 & 1.000 & 0.423 & 0.289 & 0.341 \\
				\bottomrule
		\end{tabular}}
	\end{table}
	
	\subsection{Retrieval Stability}
	\label{sec:app_retrieval}
	Attribution-conditioned retrieval builds, for each patient, one query from the top physiological features and one from the top treatment features, encodes it with MedCPT, and retrieves the top-5 documents from a fixed corpus of 3,668 PubMed abstracts (2009--2019). Table~\ref{tab:retrieval} gives the Jaccard overlap of the documents retrieved under each updated model with those retrieved under the source model, in the final era. It measures how much retrieval moves, not whether the retrieved documents are relevant.
	
	\begin{table}[!ht]
		\centering
		\caption{Retrieval stability: Jaccard similarity of top-5 MedCPT PubMed documents retrieved by updated models vs.\ the source model (3,668-abstract corpus). Final era (2020--2022), incident evaluation cohort, up to 150 positive and 150 negative cases per label; tree physiology $= 1.000$ by construction.}
		\label{tab:retrieval}
		\small
		\resizebox{\columnwidth}{!}{\begin{tabular}{llrcccc}
				\toprule
				& & & \multicolumn{2}{c}{Physiology query} & \multicolumn{2}{c}{Treatment query} \\
				Label & Model & Cases & B & C & B & C \\
				\midrule
				Vasopressor & Neural & 271 & 0.298 & 0.264 & 0.242 & 0.294 \\
				Vasopressor & Trees & 271 & 1.000 & 0.375 & 0.338 & 0.368 \\
				Intubation & Neural & 219 & 0.252 & 0.238 & 0.413 & 0.411 \\
				Intubation & Trees & 219 & 1.000 & 0.474 & 0.604 & 0.559 \\
				\bottomrule
		\end{tabular}}
	\end{table}
	
	\subsection{Selection of Excluded Stays}
	\label{sec:app_balance}
	\begin{table}[!ht]
		\centering
		\caption{Demographic and clinical balance of 2020--2022 stays excluded for lacking ICU input/procedure records versus retained stays (one row per stay; 15,086 of the 15,363 flagged stays were present in the pre-clean table). All standardized mean differences (SMD) are below 0.10 for these four variables; this does not rule out selection on unmeasured variables. Counts differ from Table~\ref{tab:audit}, which counts all raw ICU stays rather than the modelling cohort.}
		\label{tab:exclusion_bias}
		\small
		\resizebox{\columnwidth}{!}{\begin{tabular}{llccc}
				\toprule
				\textbf{Variable} & \textbf{Type} & \textbf{Excluded ($N=8{,}124$)} & \textbf{Retained ($N=6{,}962$)} & \textbf{SMD} \\
				\midrule
				Age (years) & Continuous & $65.13 \pm 16.25$ & $65.12 \pm 16.18$ & $0.0005$ \\
				Weight (kg) & Continuous & $78.50 \pm 8.92$ & $78.68 \pm 9.89$ & $-0.0191$ \\
				Female (\%) & Binary & 43.6\% & 43.3\% & $0.0069$ \\
				In-hospital Mortality (\%) & Binary & 12.2\% & 11.4\% & $0.0266$ \\
				\bottomrule
		\end{tabular}}
	\end{table}
	
	\subsection{Conceptual Architecture of the Update Arms}
	\label{sec:app_arch}
	\begin{figure}[!ht]
		\centering
		\includegraphics[width=0.4\textwidth]{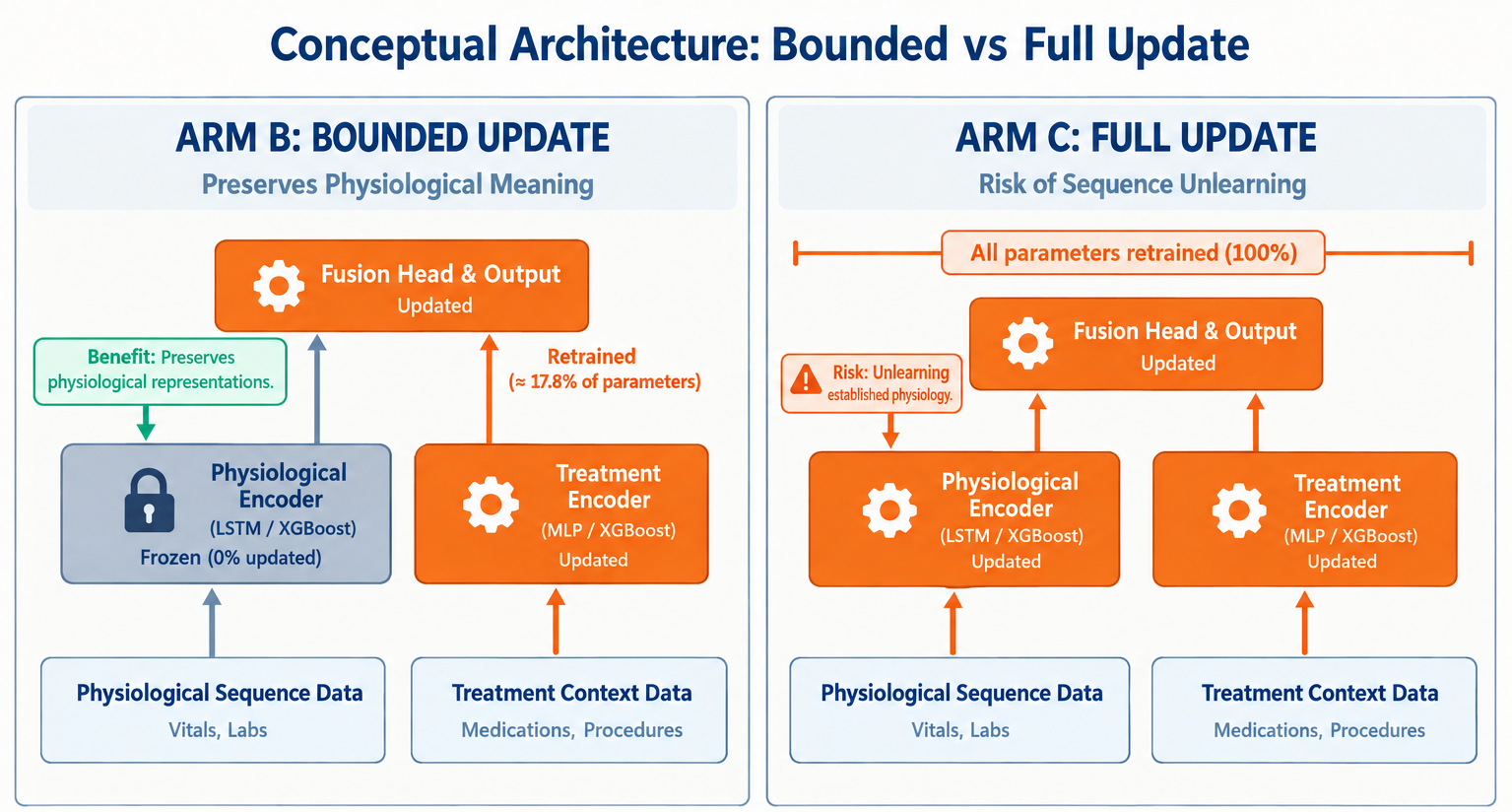}
		\caption{Conceptual architecture of the update arms. The bounded update (Arm B) freezes the physiological encoder to preserve semantic meaning, whereas the full update (Arm C) retrains all parameters, with a hypothesised risk of physiological unlearning. The treatment stream contains no eMAR features, which were excluded.}
		\label{fig:arch}
	\end{figure}
	
\end{document}